\documentclass[letterpaper, 10 pt, conference]{ieeeconf}  

\IEEEoverridecommandlockouts                              

\usepackage{graphics} 
\usepackage{epsfig} 
\usepackage{mathptmx} 
\usepackage{times} 
\usepackage{amsmath} 
\usepackage{amssymb}  
\usepackage[utf8]{inputenc}
\usepackage{array}
\usepackage{booktabs}
\usepackage{graphicx}
\usepackage{array}
\usepackage{subcaption}
\usepackage{hyperref}
\usepackage{tabularx}
\usepackage{color}
\usepackage{tikz}
\usepackage{pgfplots}
\usepackage{pgf-umlsd}
\usepackage{ifthen}
\usepackage{footnote}

\makesavenoteenv{tabular}

\newcolumntype{b}{X}
\newcolumntype{s}{>{\hsize=.5\hsize}X}

\title{\LARGE \bf
Enhanced free space detection in multiple lanes based on single CNN with scene identification*
}

\author{Fabio Pizzati$^{1}$, Fernando García$^{2}$
\thanks{*The NVIDIA Titan Xp GPU used for this research was donated by the NVIDIA
Corporation.}
\thanks{$^{1}$Fabio Pizzati is with University of Parma and University of Bologna, Italy
        {\tt\small fabio.pizzati2@unibo.it}%
}
\thanks{$^{2}$Fernando García is with Intelligent Systems Laboratory, Universidad Carlos III de Madrid, Spain
        {\tt\small fegarcia@ing.uc3m.es}}%
}

\begin{document}

\maketitle
\thispagestyle{empty}
\pagestyle{empty}

\begin{abstract}

Many systems for autonomous vehicles' navigation rely on lane detection. Traditional algorithms usually estimate only the position of the lanes on the road, but an autonomous control system may also need to know if a lane marking can be crossed or not, and what portion of space inside the lane is free from obstacles, to make safer control decisions. On the other hand, free space detection algorithms only detect navigable areas, without information about lanes. State-of-the-art algorithms use CNNs for both tasks, with significant consumption of computing resources. We propose a novel approach that estimates the free space inside each lane, with a single CNN. Additionally, adding only a small requirement concerning GPU RAM, we infer the road type, that will be useful for path planning. To achieve this result, we train a multi-task CNN. Then, we further elaborate the output of the network, to extract polygons that can be effectively used in navigation control. Finally, we provide a computationally efficient implementation, based on ROS, that can be executed in real time. Our code and trained models are available online.

\end{abstract}

\section{INTRODUCTION}

\subsection{State of the art}

Knowing the position of the lanes is essential to move the vehicle correctly on the street, and to avoid collisions with other road users. For this reason, lane detection holds great importance for assisted and autonomous driving, as ADAS for both lane departure warning and lane keeping assist \cite{narote2018review} need reliable information about lane boundaries. In their general form, lane detection algorithms address the problem using a three-step approach \cite{hillel2014recent}: in a preliminary phase, images are pre-processed, to filter noise and obstacles, and to facilitate further detections. In simple implementations, this process can be performed using only visual data, acquired by the cameras on the vehicle.

The pre-processing can be a color space transformation, often used for noise reduction and to mitigate the hard shadows impact in the detection process \cite{katramados2009real, alvarez2007shadow}. Others \cite{felisa2010robust} apply a Bird's-Eye View transform to the image. In this way, lane markings appear parallel, and with a constant width for all their length, making them easier to identify.

The second step of the typical elaboration pipeline is feature extraction. With the advent of deep learning, this process has gradually been delegated to Convolutional Neural Networks (CNNs), usually obtaining far superior performances with respect to traditional feature extraction algorithms that rely on hand-crafted kernels, but with an increasing need for labeled datasets and computing power. The CNNs for lane detection are usually trained to detect lane boundaries \cite{NEVEN2018T} or lane markings \cite{lee2017vpgnet}.

Lastly, the output of the network is post-processed and filtered, to get information about the geometric structure of the lanes. Here, some approaches use a clustering algorithm to distinguish between different lanes \cite{liu2018lane}; others fit a model for each segment of interest, usually a polynome. For instance, Neven et al. \cite{NEVEN2018T} train a CNN to separate the pixels belonging to different lane boundaries, then they fit the detected points using a third-degree polynome. Many others (\cite{wang2018lane, xu2017lane}) use the RANSAC algorithm to remove outliers. For a more insightful review of current lane detection algorithms, we refer to \cite{hillel2014recent, narote2018review}.

To correctly position a vehicle inside a lane, information regarding road areas is likewise needed, as lanes can be either free or occupied by vehicles or other obstacles. In literature, many approaches exploit visual and LIDAR data jointly with geometric properties of the scene to filter obstacles and detect interesting areas \cite{liu2018co}. Others \cite{liu2018segmentation, sanberg2017free} use a CNN, training it on labeled images where the road is annotated. In a recent work, Caltagirone et al. \cite{caltagirone2019lidar} developed a data fusion strategy to use either image and LIDAR data in a deep learning based algorithm, and obtain state-of-the-art performances on the KITTI dataset benchmarks for road detection.

Modern deep learning techniques are proven to solve the road detection task efficiently, at the point it can be easily coupled with other tasks. An example can be found in \cite{Teichmann2018MultiNetRJ}, where a single CNN is trained for street classification, road areas detection and vehicle detection.

\subsection{Motivations}

\begin{figure*}[ht]
  \caption{System overview}\label{figure:system_overview}
	\noindent\resizebox{\textwidth}{!}{
		\includegraphics[width=\linewidth]{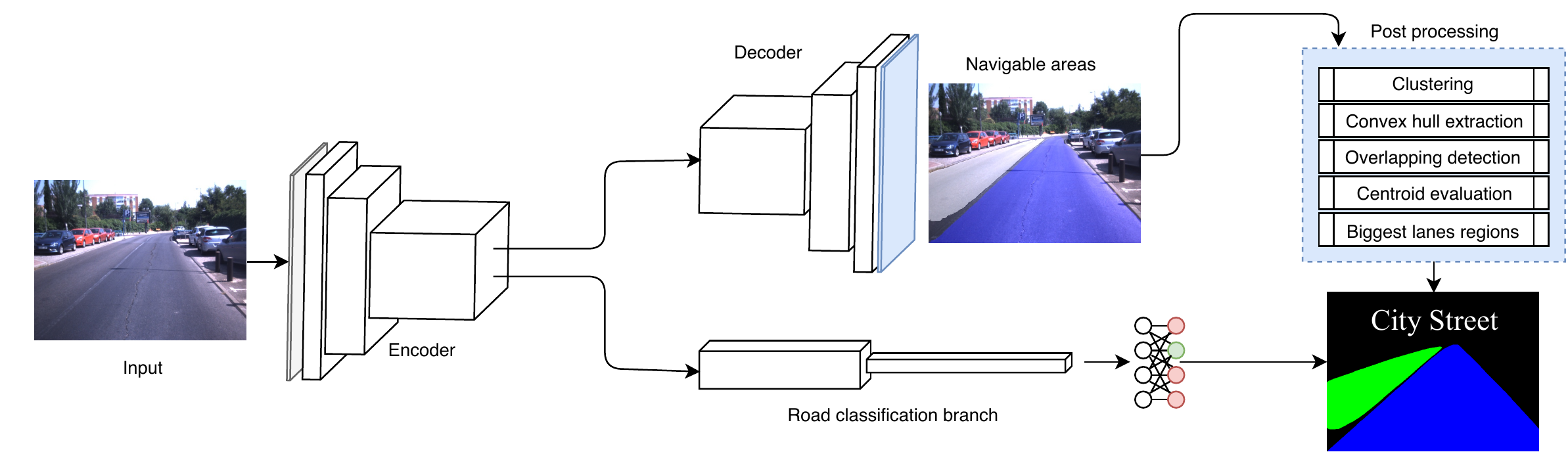}}
  \footnotesize\textbf{Details:} the input image is firstly propagated into the encoder, then in the two branches, the decoder for the detection of drivable areas and the road classification branch to classify the road type, where fully connected layers are added to the end. After the post-processing applied to the output of the decoder, we detect the drivable areas, divided by lane, and the road class.
\end{figure*}
In lane detection, training the CNN on lane boundaries or markings ease the learning procedure, as they are the most noticeable feature used to distinguish among different lanes even for human agents. This training strategy, unfortunately, leads the neural network to highly depend on the lane markings, that are often unavailable, especially in country or urban roads. Besides this, empirical results \cite{DBLP:journals/corr/abs-1806-05525} demonstrated that the classical supervised approach detects thicker lane boundaries with respect to the ones in the ground truth, being the class ratio in a typical road environment heavily unbalanced. Moreover, as already highlighted, to plan the path of a vehicle correctly, the obstacles positions and, more generally, the navigable areas, are needed as well. For this reason, it is necessary to use other CNNs for obstacle or free space detection, but GPU intensive operations need to be reduced to a minimum on an autonomous vehicle, considering limited memory availability and power supply. Lastly, detecting lanes or road areas does not give information about the traffic direction in them, so, for example during a lane change, if no further distinctions are made, the vehicle may risk a frontal collision with others, incoming in the opposite direction. To the best of our knowledge, there are no publicly available datasets with this kind of information.

\subsection{Proposed approach}

We propose an alternative approach: we directly detect all the pixels belonging to the road area in each lane, with a single CNN. The CNN we introduce is also used to classify the street type, and this enables additional inferences about the lanes availability for navigation. We then post-process the detected pixels to extract polygons, which will be used by a path planning algorithm. In particular, we are interested in the biggest road areas in the ego lane and the lateral lanes, if existing. We implemented the system using the ROS framework. In this way, we introduced a method that manages several tasks that are typically solved with the usage of multiple CNNs and data fusion strategies, significantly reducing the computing power required.

The paper is organized as follows: in section II we introduce the algorithm and its components, while in section III we provide technical details about the training strategy and our ROS implementation. Finally, in section IV we present the results obtained on the dataset we used and the qualitative results on our sequences; section V concludes the paper. For the rest of the paper, we refer to road areas free from obstacles as \textit{drivable areas}.

\section{METHOD}

\subsection{Algorithm description}

The algorithm is based on a CNN for drivable areas detection. During the training phase, we differentiate between the \textit{ego lane} areas, defined as all the space included in the lane the vehicle is driving in, and the \textit{other lanes} areas. Other lanes are all labeled with the same class in our training set, as further distinctions between different lanes are performed in a subsequent step. Additionally, we use the same neural network to classify the street type: this information can be exploited to better understand if the detected space can be used or not. For example, if the vehicle travels on a highway, it is evident that all the detected space may be used, as there are no lanes reserved to the opposite traffic, but this may not be true in an urban scenario. A detailed description of this is given in section \ref{exploiting}.

We then post-process the output of the CNN, clustering the detected points for the two different classes and extracting the convex hull for each of them. Each cluster represents a drivable area and belongs to a lane. With this process, functional information for a planning algorithm is extracted, but overlapping areas, assigned to two different polygons simultaneously, can be found. To solve the problem, we assign the overlapping area to one polygon. We calculate the coordinates of the centroid for each extracted polygon, so we can easily differentiate between right and left lanes drivable areas. The polygons with the greatest area, both in the ego lane, the left lane, and the right lane, are our desired output, as they represent the free areas for each lane, and so the most valuable information for navigation in the scene. Technical details are provided in section \ref{technical-pp}, while the system is presented in figure \ref{figure:system_overview}.

\subsection{The CNN architecture}
\begin{table}
  \caption{\upshape{CNN Architecure}}
  \setlength{\tabcolsep}{0.55\tabcolsep}
  \begin{subtable}[t]{\linewidth}
  \caption{Panel A: Encoder}
  \centering
  \begin{tabular}{ *{7}{c} }
    \toprule
    \textbf{Layer} & \textbf{Type} & \textbf{Kernel} & \textbf{Stride} & \textbf{Dilation} & \textbf{Dropout} & \textbf{Output} \\
    \midrule
    1 & Downsamp & -  & -    & - & - & \(16\times 320 \times 240\) \\
    2 & Downsamp & -  & -    & - & - & \(64\times 160 \times 120\) \\
    3-7 & Non-bt-1d & -  & -    & 1 & 0.03 & \(64\times 160 \times 120\) \\
    8 & Downsamp & -  & -    & - & - & \(128\times 80 \times 60\) \\
    9 & Non-bt-1d  & - & -    & 2 & 0.3 & \(128\times 80 \times 60\) \\
    10 & Non-bt-1d & -  & -    & 4 & 0.3 & \(128\times 80 \times 60\) \\
    11 & Non-bt-1d & -  & -    & 8 & 0.3 & \(128\times 80 \times 60\) \\
    12 & Non-bt-1d & -  & -    & 16 & 0.3 & \(128\times 80 \times 60\) \\
    13 & Non-bt-1d & -  & -    & 2 & 0.3 & \(128\times 80 \times 60\) \\
    14 & Non-bt-1d & -  & -    & 4 & 0.3 & \(128\times 80 \times 60\) \\
    15 & Non-bt-1d & -  & -    & 8 & 0.3 & \(128\times 80 \times 60\) \\
    16 & Non-bt-1d & -  & -    & 16 & 0.3 & \(128\times 80 \times 60\) \\
    \bottomrule
  \end{tabular}
  \vspace{1pt}
\end{subtable}
\begin{subtable}[t]{\linewidth}
  \vspace{6px}

  \caption{Panel B: Decoder branch}

\begin{tabular}{ *{7}{c} }
  \toprule
  \textbf{Layer} & \textbf{Type} & \textbf{Kernel} & \textbf{Stride} & \textbf{Dilation} & \textbf{Dropout} & \textbf{Output} \\
  \midrule
  1 & Upsamp  & - & -    & - & - & \(64\times 160 \times 120\) \\
  2-3 & Non-bt-1d  & - & -    & 1 & 0 & \(64\times 160 \times 120\) \\
  4 & Upsamp  & - & -    & - & - & \(16\times 320 \times 240\) \\
  5-6 & Non-bt-1d  & - & -    & 1 & 0 & \(16\times 320 \times 240\) \\
  7 & Deconv     & \(2\times2\) & 2      & - & - & \(3\times 640 \times 480\) \\
  \bottomrule
\end{tabular}
\end{subtable}
\begin{subtable}[t]{\linewidth}
  \vspace{6px}
  \caption{Panel C: Road classification branch}

\begin{tabular}{ *{7}{c} }
  \toprule
  \textbf{Layer} & \textbf{Type} & \textbf{Kernel} & \textbf{Stride} & \textbf{Dilation} & \textbf{Dropout} & \textbf{Output} \\
  \midrule
  1 & Conv2d & \(3\times3\)  & 2   & 1 & - & \(256\times 40 \times 30\) \\
  2 & MaxPool2d   & \(2\times2\)  & 2 & - & - & \(256\times 20 \times 15\) \\
  3 & Non-bt-1d   & -    & - & 1 & 0.3 & \(256\times 20 \times 15\) \\
  4 & Conv2d & \(3\times3\)  & 2   & 1 & - & \(512\times 10 \times 8\) \\
  5 & MaxPool2d   & \(2\times2\)  & 2 & - & - & \(512\times 5 \times 4\) \\
  6 & Non-bt-1d   & -    & - & 1 & 0.3 & \(512\times 5 \times 4\) \\
  7 & Linear      & -    & - & - & - & 1024 \\
  8 & Linear      & -    & - & - & - & 4 \\
  \bottomrule
\end{tabular}
\end{subtable}
\vspace{0.5em}

\footnotesize{\textbf{Notes:} the features extracted by the encoder are propagated in the two branches simultaneously. Additional layers like PReLU non-linearities and Batch Normalization layers are not listed. When a layer parameter is not available, it is listed as -. The output size is calcluated for a \(640\times480\) input.}
\label{architecure}
\end{table}

As already mentioned, many approaches in literature for lane and free space detection use a fully convolutional network \cite{long2015fully}, that is, by definition, a structure entirely composed by convolutional layers. This permits, in a typical use case scenario, to assign a class to each pixel of an image.
The CNN architecture that we chose to train is ERFNet \cite{romera2018erfnet}, a fully convolutional network that holds the best position in terms of \(\frac{mIoU}{frames\_per\_second}\) ratio in the Cityscapes dataset benchmarks for semantic segmentation \cite{cordts2016cityscapes}. The mIoU is defined as the mean of the single intersection over union per class, that is:

\begin{equation}
IoU = \frac{TP}{FP + TP + FN}
\end{equation}
where TP, FP, and FN denote, respectively, the true positives, the false positives, and the false negatives for a single class. ERFNet is composed by an \textit{encoder} and a \textit{decoder}, where the first maps the pixels of the image to a feature vector of a given size, while the other extracts a representation of the initial scene in a different domain, usually the chosen pixel-wise classification. The essential components of the network are the \textit{non-bottleneck-1d} residual block, the \textit{downsampler} block and the \textit{upsampler} block. The residual block is composed by two couples of asymmetric \(1\times3\) and \(3\times1\) convolutions. The other two blocks use pooling layers and transposed convolutions, respectively, to reduce or increase the feature map size. Further details can be found in the original implementation \cite{romera2018erfnet}.

To be able to classify the road type, the ERFNet architecture has been modified adding additional layers in a separate branch. During the inference phase, the extracted feature vector is propagated in the newly defined layers and in the decoder, and the two branches are trained with the corresponding data. With this strategy, our CNN can simultaneously detect free areas on a pixel level, and classify the whole image. The full architecture of our network is given in table \ref{architecure}. The CNN has been implemented and trained using the PyTorch framework.

\subsection{Dataset}

\begin{figure}[h]
  \caption{BDD100K sample images and drivable areas labels}\label{figure:bdd_samples}
\begin{subfigure}{\linewidth}
  \begin{subfigure}{0.49\linewidth}
    \includegraphics[width=\textwidth]{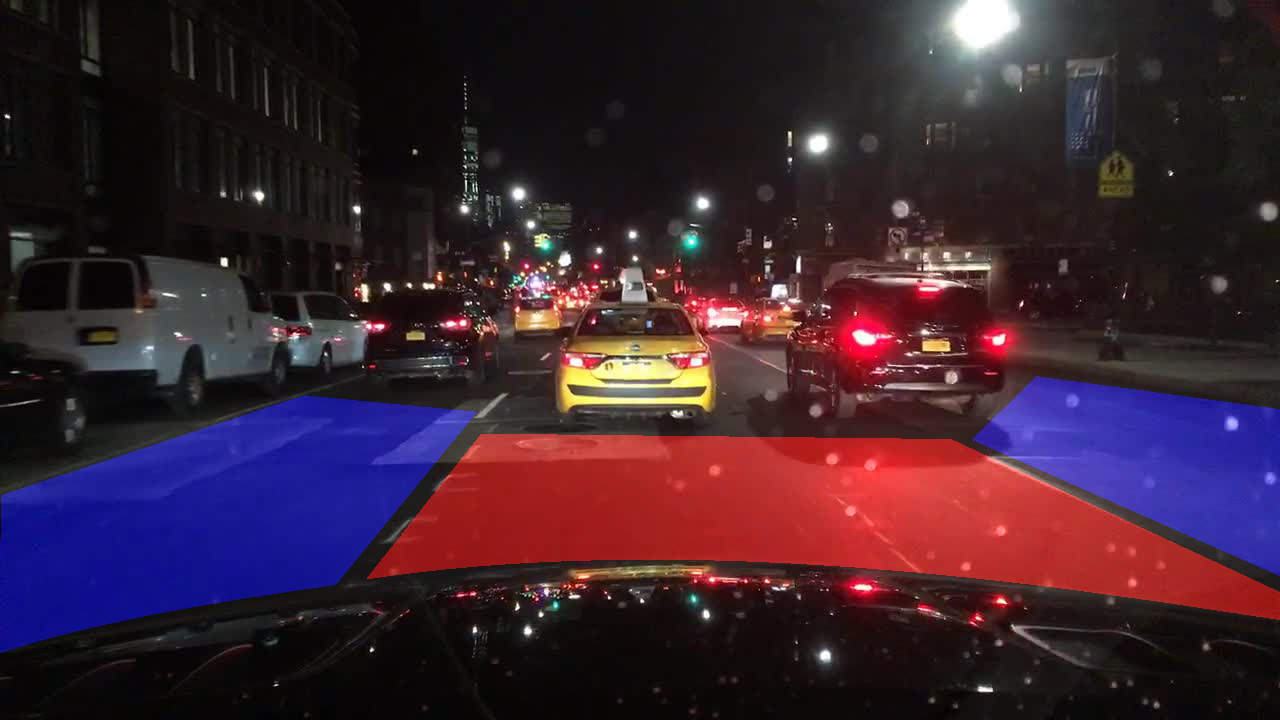}
  \end{subfigure}
  \hfill
  \begin{subfigure}{0.49\linewidth}
    \includegraphics[width=\textwidth]{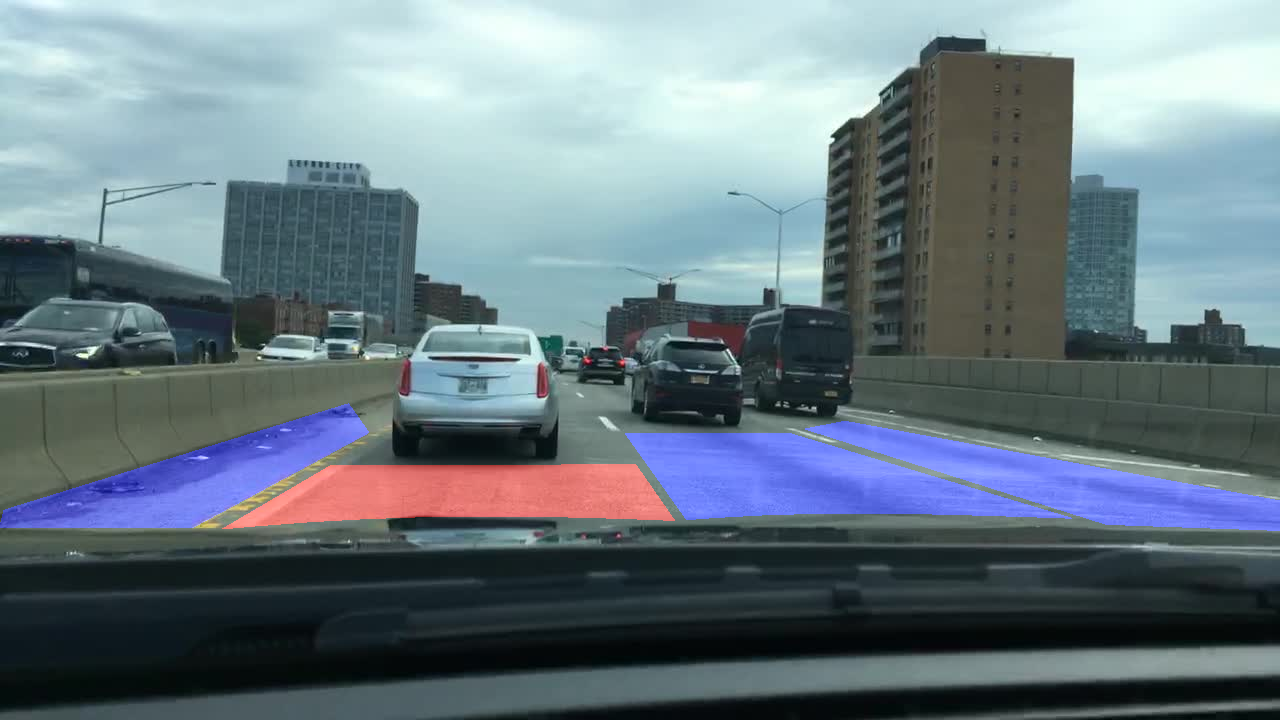}
  \end{subfigure}
  \hfill
  \vspace{6px}
\end{subfigure}
\footnotesize
\textbf{Details:} The red color corresponds to the ego lane areas, while the other lanes areas are painted as blue.
\end{figure}

As in all the deep learning based approaches, data quality and quantity is fundamental to achieve good results. Given that our network has two different branches, both pixel-wise annotations regarding drivable areas inside lanes and street type classification are needed for each image. For this reason, we chose to train the network on the Berkeley DeepDrive dataset \cite{yu2018bdd100k} (BDD), which is composed by 100k \(1280\times720\) images, divided in 70k for the training set, 10k for the validation set and 20k for the test set. It includes, as needed, annotations for drivable areas, and image-level descriptive labels containing, among the others, the road type. In particular, for each pixel of drivable areas, the class \textit{ego lane}, \textit{other lanes} or \textit{background} can be assigned. The \textit{other lanes} class includes all the pixels that do not belong to the ego lane but are still part of the road. As regards the image classification task, 7 classes are used as labels for the road type: \textit{residential}, \textit{highway}, \textit{city street}, \textit{parking lot}, \textit{gas station}, \textit{tunnel} and \textit{undefined}. We decided to use only four of them, assigning the images belonging to the classes \textit{parking lot}, \textit{gas stations}, \textit{tunnel} and \textit{undefined} to a single class, called \textit{others}. Examples of images and annotations are shown in figure \ref{figure:bdd_samples}.

\subsection{Exploiting the road type} \label{exploiting}
The capability of our network to classify the road type eases decisions regarding whether to use or not the navigable area in the other lanes. Sampling several hundreds of images from the BDD100K dataset and qualitatively evaluating them, we noticed that several behavioral rules could be extracted.
\subsubsection{Highway scenarios} In our sampling of the dataset, all frames labeled as \textit{highway} are one-way streets with alternate lanes (we define those as \textit{multi-lane} streets). Therefore, the vehicle may perform a lane change and use the other lanes areas, if needed.
\subsubsection{Residential scenarios} Two-way or single-lane streets are labeled as \textit{residential}, so, even if a side lane is detected, the vehicle should remain in the ego lane.
\subsubsection{Others scenarios}
In the \textit{others} class, we grouped the three street categories where the proposed task is most challenging, so, in the case this scenario is detected, the vehicle may rely only on the drivable space of the ego lane for navigation, to minimize the number of errors.
\subsubsection{City Street scenarios}
Lastly, if a \textit{city street} is detected, there won't be enough information to perform a lane change, as in the BDD dataset both two-way and multi-lane frames are labeled as \textit{city street}.
\section{Technical details}

\subsection{Multi-task CNN training}
The first phase of our algorithm employs a multi-task network, where one task is free space segmentation, and the other road classification. Recent studies by Kendall et al. \cite{kendall2017multi} demonstrated how uncertainty in neural networks could be exploited to achieve better predictions in multi-task networks. Following their hypothesis, we assume that there is an uncertainty that depends on the task we are performing, but not on the data, called \textit{homoscedastic uncertainty}. We quantify the uncertainty for each task introducing two variables, \(\sigma_{fs}\) for the uncertainty associated with the drivable areas segmentation task, and \(\sigma_{c}\) for the one associated with the road classification task. These uncertainties are constant, so it is reasonable to assume that we can estimate them within the optimization process, jointly with the neural network parameters. We are going to use these quantities as weights for the different loss functions the CNN needs.
For the two tasks, two weighted cross-entropy losses are used, each one defined as:

\begin{equation}
\label{Loss}
L_{ce} = - \sum_{c = 0}^M w_cy_{p,c}log(p_{p,c})
\end{equation}

In \ref{Loss}, \(M\) is the number of classes, \(w_c\) is the weight for class \(c\), \(y_{p,c}\) is \(1\) if the prediction \(p\) has class label \(c\) and 0 otherwise, and \(p_{p,c}\) is the probability estimated by the CNN for the prediction \(p\) to have as label \(c\). The weights for the free space detection task are all set as 1. For the road classification task, being the dataset heavily unbalanced against the \textit{others} class, we weighted the classes using the strategy defined in \cite{paszke2016enet}:

\begin{equation}
\label{weights}
w_{c} = \frac{1}{log(k + p_c)}
\end{equation}
In \ref{weights}, \(p_c\) is the probability of an element to have label \(c\), and \(k\) is a regularization value set at \(1.02\).

We define as \(L_{fs}\) the bidimensional cross-entropy loss associated with the drivable areas detection, and as \(L_c\) the cross entropy loss for the road type classification. During the backward step, we use as loss function the weighted sum of the two losses:

\begin{equation}
L = \frac{1}{\sigma_{fs}^2}L_{fs} + \frac{1}{\sigma_{c}^2}L_c + log(\sigma_{fs}) + log(\sigma_{c})
\end{equation}

We trained the network using the Adam optimizer, for 80 epochs on the BDD training set downsampled at \(640\times360\) resolution, using a starting learning rate of \(10^{-4}\) and weight decay \(10^{-4}\). Instead of \(\sigma_{cs}\) and \(\sigma_{c}\), \(log(\sigma_{cs})\) and \(log(\sigma_{c})\) are estimated, in order to achieve better training stability. The two \(log(\sigma)\) values are initialized at \(0\), and modified at each iteration. We use a polynomial learning rate decay for each epoch as in \cite{romera2018erfnet}, with exponent set to \(0.9\). Data augmentation has been applied to get better performances, in the form of random horizontal flipping with probability 0.5, and random translation, both for \(x\) and \(y\) axes, in the range of \(\{-2,2\}\) pixels.

\subsection{Post-processing} \label{technical-pp}
\begin{figure}[t]
  \caption{Convex hulls intersection}
  \includegraphics[width=\linewidth]{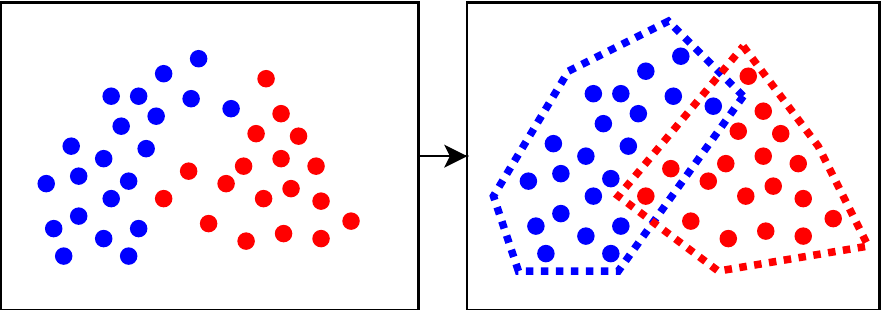}
  \footnotesize{\textbf{Details:} Blue and red points belong to two different clusters. The convex hull extraction generates an intersecting region that, in our algorithm, is never assigned to an ego lane region.
  }
  \label{chull}
\end{figure}

The post-processing algorithm is necessary to transform the pixel-wise free space detection to a set of polygons that will be effectively used in navigation control. In a first step, the output of the neural network is downsampled by 4, to speed up further processing and maintain real-time performances. Then, the points are grouped using a density-based clustering with the DBSCAN algorithm, separately on the points of the two classes, to parallelize the operation. Once this phase is terminated, the convex hull is extracted for each cluster, obtaining a first set of drivable regions. This process may generate overlapping areas between different regions, so a filtering step is required, where the intersections between two different polygons are deleted from one of them. An example of this behavior is explained in figure \ref{chull}. In this case, if a drivable region belongs to the ego lane, we remove the intersection region from it, to minimize the probability to invade another lane. Instead, if the ego lane has no overlapping regions, the area is removed from a randomly chosen polygon between the two intersecting. In our experiments, this scenario has never occurred.

In a final step, we separate the drivable regions on different sides of the ego lane, comparing the \(x\) coordinate of the centroid of the drivable areas of the side lanes with the \(x\) coordinate of the centroid of the biggest polygon of the ego lane. The output of the algorithm gives us the biggest region of the ego lane and the lateral lanes, if existing.

\subsection{ROS Implementation}

To integrate the proposed system on an autonomous vehicle, we provide a ROS implementation. This framework makes it possible to quickly implement pipelines between different nodes, and enables communication with others on-board processing systems in a simple way. We designed two ROS nodes: the first of them receives as input the images acquired by the camera, and process them with the CNN. Then, it sends to the second node the output of the two branches. The second node, instead, executes the post-processing algorithm on the points of the drivable areas and outputs the desired drivable regions with the road type. In a future implementation, this output will be forwarded to the path planning algorithm. In the second node, to further speed up the whole process, a \texttt{threadPool} structure is implemented, that enables multithreading for clustering jobs. Given that the clustering should be performed only on points belonging to the same class, pixels classified as \textit{ego lane} and \textit{other lanes} are clustered in two separate threads, in parallel. The initial number of available threads is fixed to 6.

Those design choices have been performed to pipeline the two most intensive operations, namely the CNN elaboration and the clustering algorithm, and to parallelize operations that don't depend on each other. These optimizations make it possible to reach an elaboration speed of over 20fps for \(640\times480\) images, using a NVIDIA Titan Xp GPU and an Intel Core i9-7900X CPU. Our code and models are available at \href{https://github.com/fabvio/ld-lsi/}{\texttt{https://github.com/fabvio/ld-lsi/}}.

\section{Results}

\subsection{Quantitative evaluation}
\begin{figure*}[h!]
\caption{Qualitative evaluations on our sequences}
\label{results_images}
\begin{subfigure}{\linewidth}
  \begin{subfigure}{0.17\linewidth}
    \includegraphics[width=\textwidth]{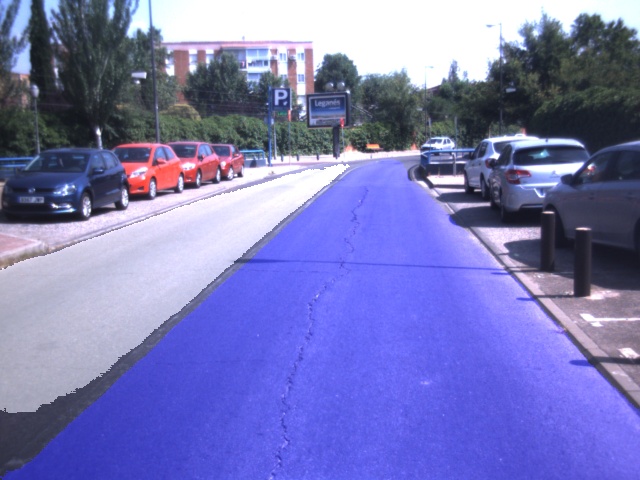}
  \end{subfigure}
  \hfill
  \begin{subfigure}{0.17\linewidth}
    \includegraphics[width=\textwidth]{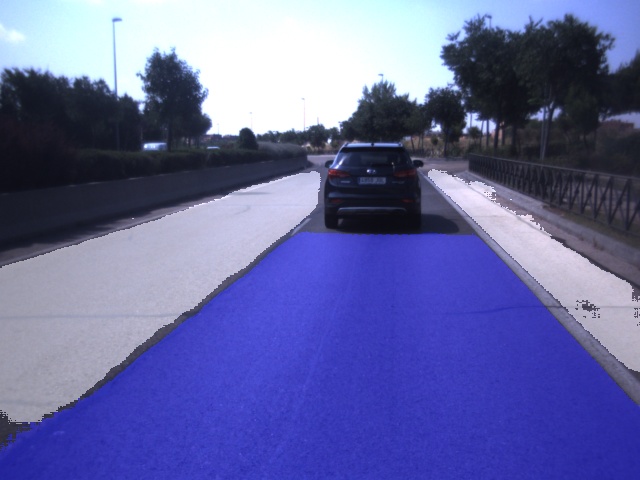}
  \end{subfigure}
  \hfill
  \begin{subfigure}{0.17\linewidth}
    \includegraphics[width=\textwidth]{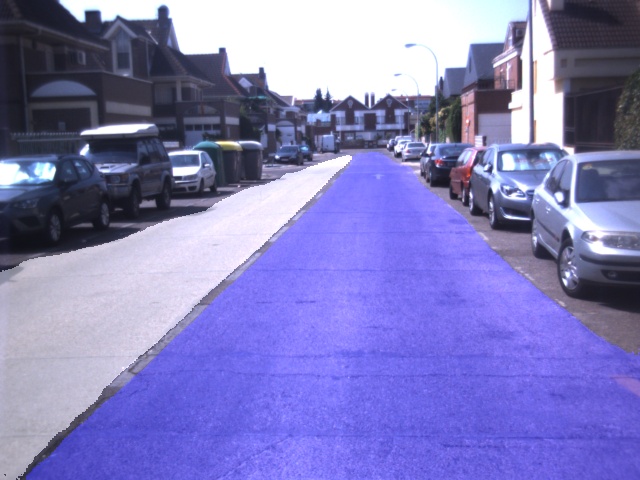}
  \end{subfigure}
  \hfill
  \begin{subfigure}{0.17\linewidth}
    \includegraphics[width=\textwidth]{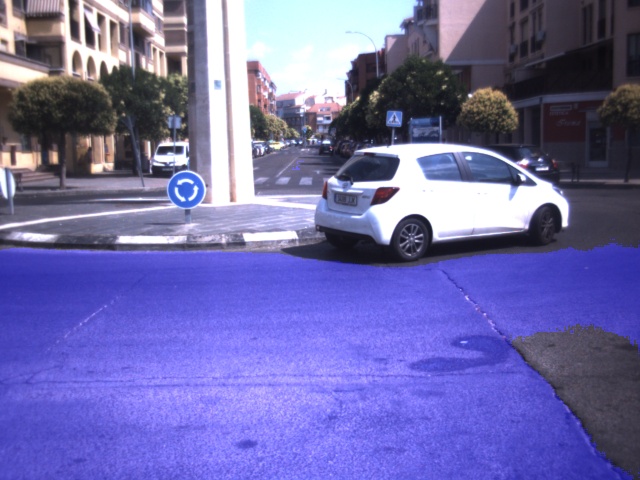}
  \end{subfigure}
  \hfill
  \begin{subfigure}{0.17\linewidth}
    \includegraphics[width=\textwidth]{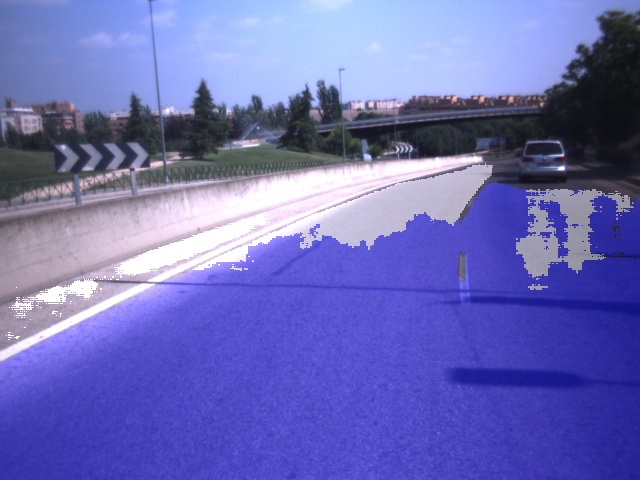}
  \end{subfigure}
  \hfill
\end{subfigure}
\begin{subfigure}{\linewidth}
  \begin{subfigure}{0.17\linewidth}
    \includegraphics[width=\textwidth]{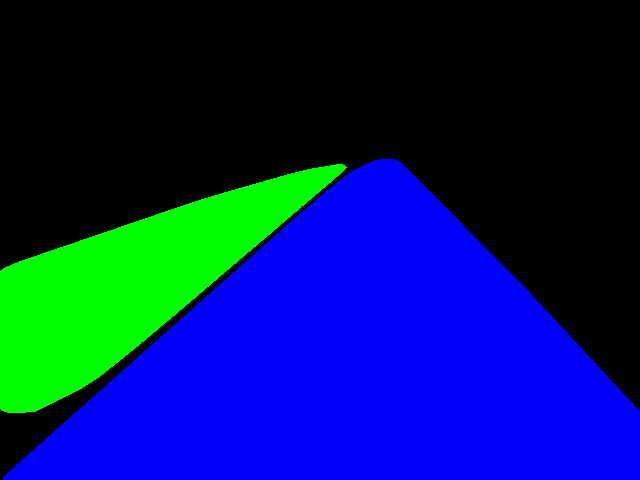}
  \end{subfigure}
  \hfill
  \begin{subfigure}{0.17\linewidth}
    \includegraphics[width=\textwidth]{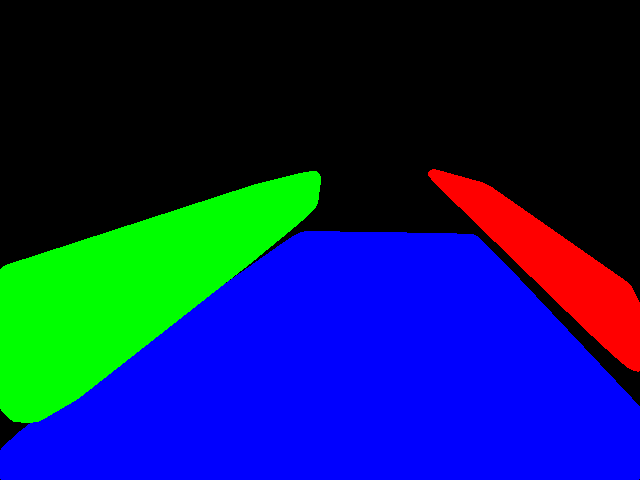}
  \end{subfigure}
  \hfill
  \begin{subfigure}{0.17\linewidth}
    \includegraphics[width=\textwidth]{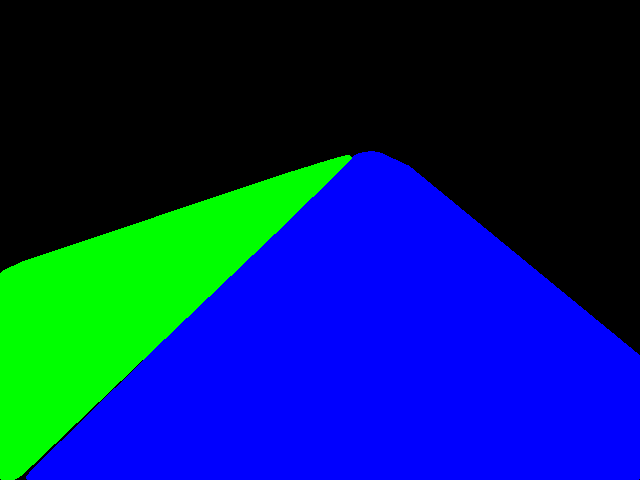}
  \end{subfigure}
  \hfill
  \begin{subfigure}{0.17\linewidth}
    \includegraphics[width=\textwidth]{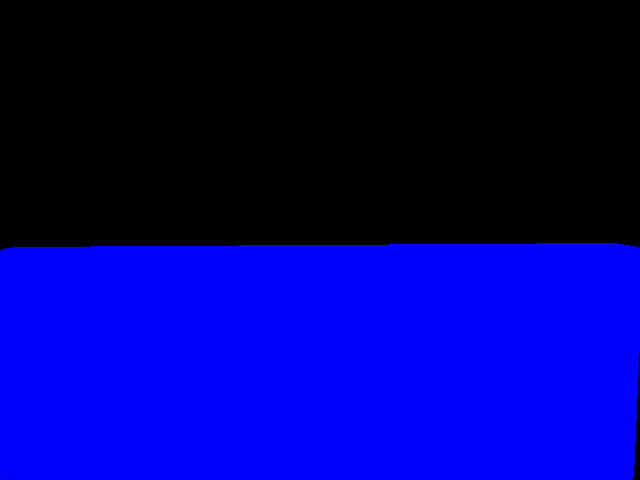}
  \end{subfigure}
  \hfill
  \begin{subfigure}{0.17\linewidth}
    \includegraphics[width=\textwidth]{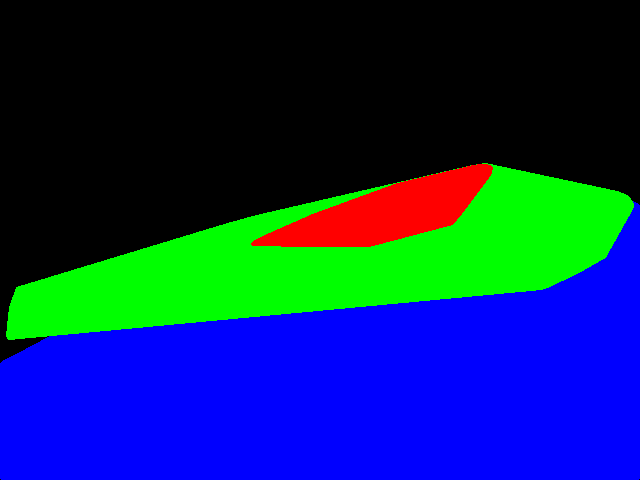}
  \end{subfigure}
  \hfill
\end{subfigure}
\begin{subfigure}{\linewidth}
  \begin{subfigure}{0.17\linewidth}
    \includegraphics[width=\textwidth]{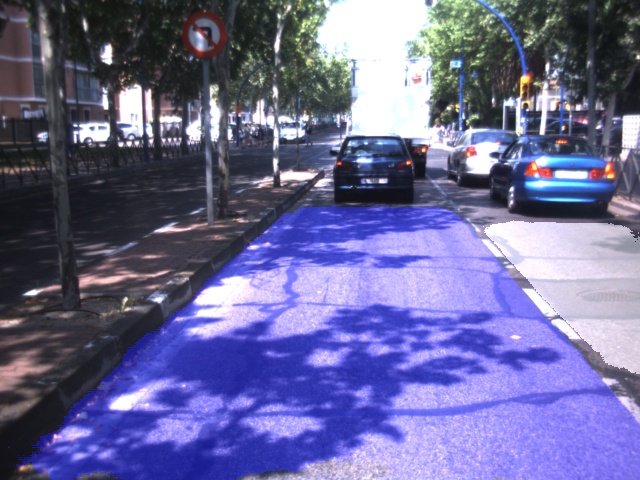}
  \end{subfigure}
  \hfill
  \begin{subfigure}{0.17\linewidth}
    \includegraphics[width=\textwidth]{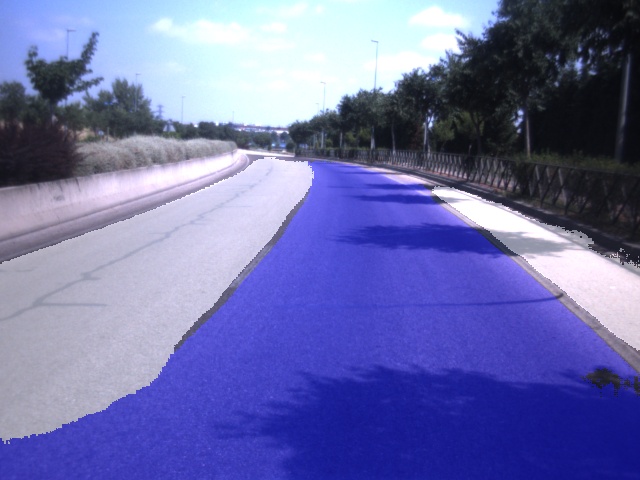}
  \end{subfigure}
  \hfill
  \begin{subfigure}{0.17\linewidth}
    \includegraphics[width=\textwidth]{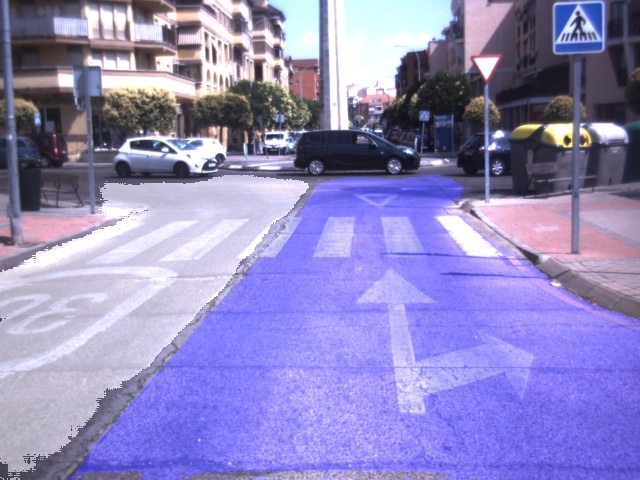}
  \end{subfigure}
  \hfill
  \begin{subfigure}{0.17\linewidth}
    \includegraphics[width=\textwidth]{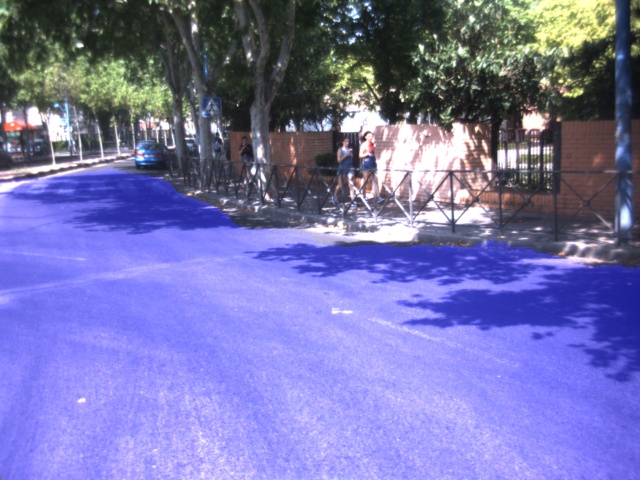}
  \end{subfigure}
  \hfill
  \begin{subfigure}{0.17\linewidth}
    \includegraphics[width=\textwidth]{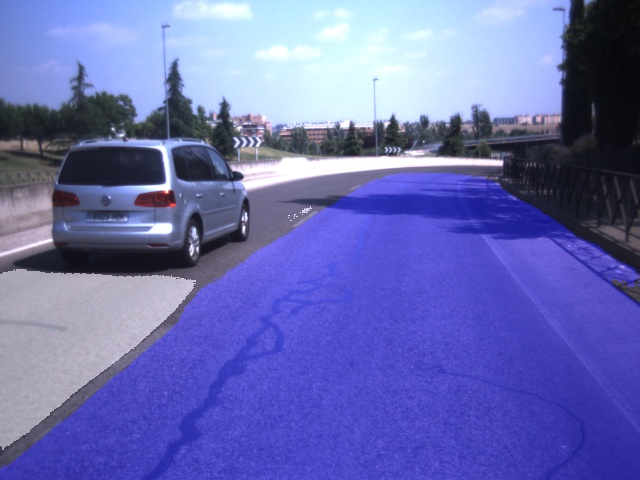}
  \end{subfigure}
  \hfill
\end{subfigure}
\begin{subfigure}{\linewidth}
  \begin{subfigure}{0.17\linewidth}
    \includegraphics[width=\textwidth]{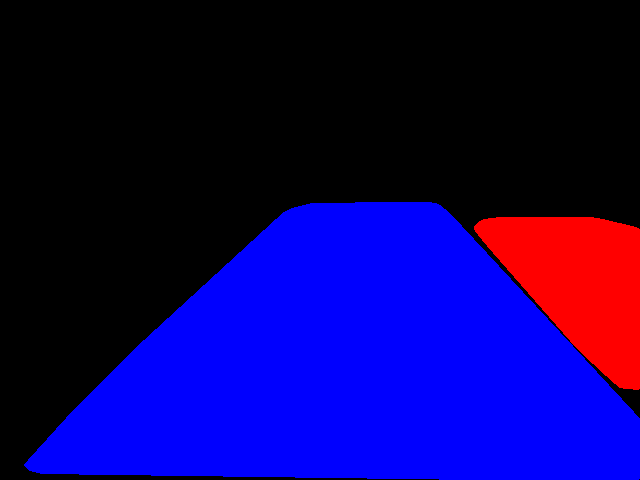}
  \end{subfigure}
  \hfill
  \begin{subfigure}{0.17\linewidth}
    \includegraphics[width=\textwidth]{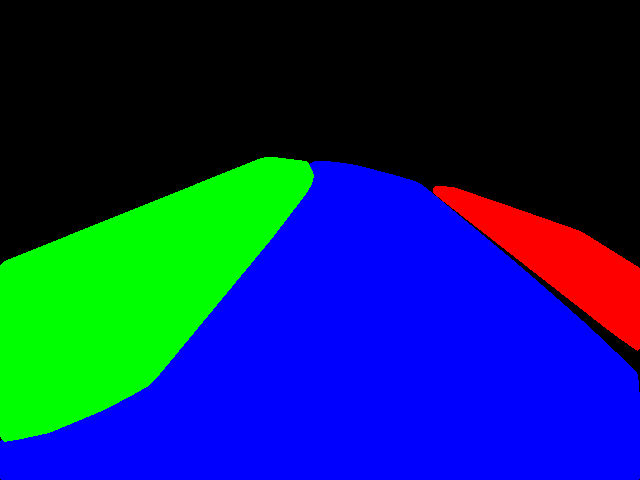}
  \end{subfigure}
  \hfill
  \begin{subfigure}{0.17\linewidth}
    \includegraphics[width=\textwidth]{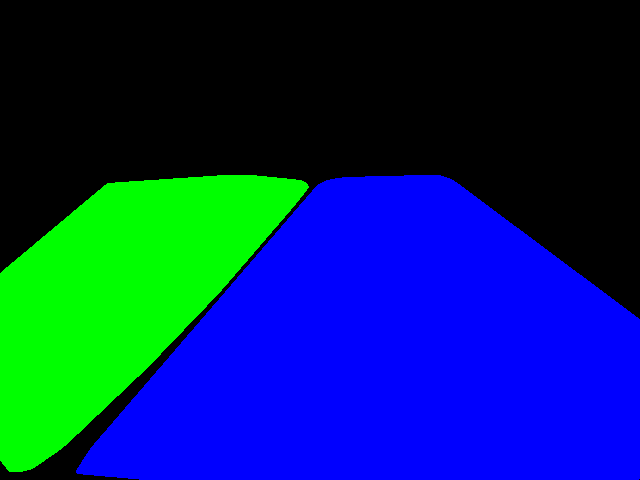}
  \end{subfigure}
  \hfill
  \begin{subfigure}{0.17\linewidth}
    \includegraphics[width=\textwidth]{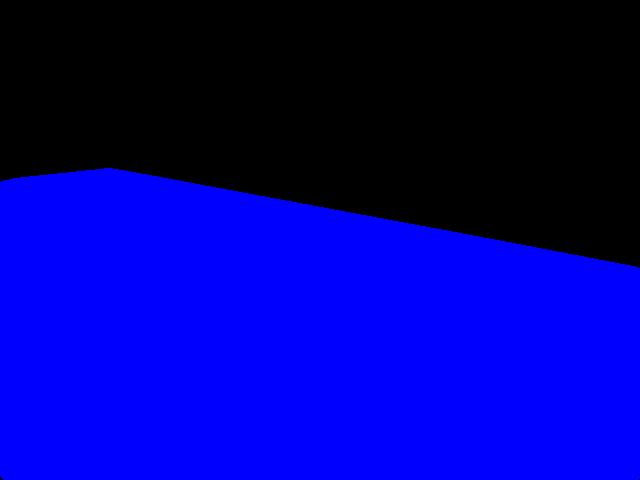}
  \end{subfigure}
  \hfill
  \begin{subfigure}{0.17\linewidth}
    \includegraphics[width=\textwidth]{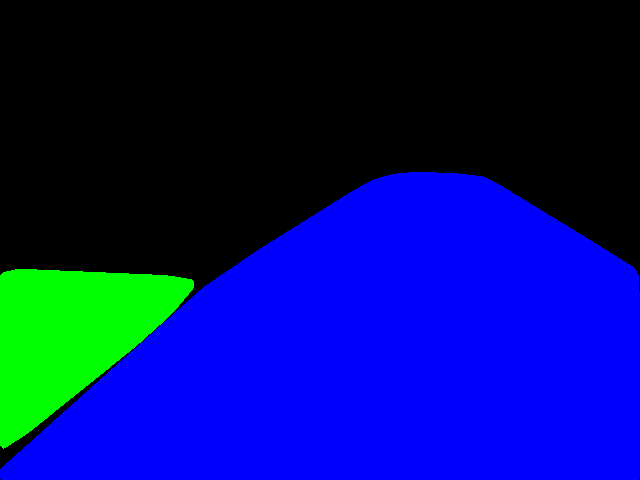}
  \end{subfigure}
  \hfill
\end{subfigure}

\par\vskip \abovecaptionskip
\begin{subfigure}{\linewidth}
  \setcounter{subfigure}{0}%
  \begin{subfigure}{0.19\linewidth}
    \caption{City street}
  \end{subfigure}\hfill
  \begin{subfigure}{0.19\linewidth}
    \caption{Highway}
  \end{subfigure}\hfill
  \begin{subfigure}{0.19\linewidth}
    \caption{Residential}
  \end{subfigure}
  \begin{subfigure}{0.19\linewidth}
    \caption{Other}
  \end{subfigure}\hfill
  \begin{subfigure}{0.19\linewidth}
    \caption{Failures}
  \end{subfigure}\hfill
\end{subfigure}
\footnotesize \textbf{Details:} the results are divided by the road type detected by the CNN. Both the CNN output for navigable areas and the extracted polygons are presented. In the CNN output, the ego lane pixels are blue, while the other lanes ones are white. For the drivable regions, the right lane is red, while the left one is green. For clarity, the ego lane remains blue. The detection of drivable areas is entirely consistent even having strong shadows and imperfections on the road (columns a, b and c). Images labeled as \textit{others} (columns d) occur more often in roundabouts, supposedly because of the absence of road markings and for the more comprehensive free area. In the last columnm, some detection failures are presented: the first one from the top is caused by a lane change, where the neural network cannot correctly identify what the ego lane is. In the second example, the emergency lane on the right is misclassified.

\end{figure*}

We evaluated the performances of our system in two ways. First, we used the dataset validation and test data to compare the output of our CNN, without post-processing, with other CNNs, trained on the same data. Differently from other systems, we had to use two different metrics, as our CNN is a multi-task network. To evaluate the quality of our navigable area detection, we use the mean intersection over union metric (\textit{mIoU}). Instead, we evaluated the road classification task in terms of accuracy, calculating the ratio between the correctly identified roads over the total number of images in our evaluation set. Here, only the performances on the validation set will be evaluated, as neither the test set labels nor an evaluation server is available. Results are presented in table \ref{results}. We compared our method with the winners of the Workshop on Autonomous Driving in CVPR 2018 challenge for drivable area detection. Please note that the networks we are comparing our algorithm with are not suitable for real-time data processing, while ours is. In particular, IBN-PSN is based on IBN-Net \cite{pan2018two}, while both Mapillary Research and DiDi AI Labs use a modified version of ResNet \cite{he2016deep} as the backbone for feature extraction. From the results, it is possible to infer that our network achieves comparable results in term of mIoU with the other approaches, and it is extremely faster, requiring only a fraction of the GMACs for inference. With the proposed training strategy, it is possible to add a branch in the baseline network, without a significant loss of accuracy with respect to the single task networks.



\begin{table}
  \caption{Quantitative evaluation on the BDD dataset}
  \label{results}
  \begin{tabularx}{\linewidth}{bsssssss}
    \toprule
    \textbf{Method} & \textbf{Scene} & \textbf{Road} & \textbf{Weight} & \textbf{mIoU\%} & \textbf{Acc.\%} & \textbf{FPS} & \textbf{GMAC}\\
    \midrule
    \centering
    IBN\_PSA/P &  & \checkmark &  & 86.18 & - & 3.81 & 327.51\\
    Mapillary &  & \checkmark &  & 86.04 & - & 0.153 & 758.37\\
    DiDiLabs & & \checkmark &  &  84.01 & - & 3.35 & 103.96\\
    Ours &  & \checkmark &  & 83.35 & - & 23.86 & 15.11\\
    Ours &\checkmark  &  &  & - & 77.73 & 27.58  & 13.51 \\
    Ours & \checkmark & \checkmark &  & 74.81 & 63.88 & 22.59 & 15.91\\
    Ours & \checkmark & \checkmark & \checkmark & \textbf{82.62} & \textbf{76.53} & \textbf{22.59} & \textbf{15.91}\\
    \bottomrule
  \end{tabularx}
  \vspace{0.5em}

  \footnotesize{\textbf{Notes:} In the table above, \textit{Scene} refers to the street classification task, while \textit{Road} is the drivable area detection task. With \textit{Weight} we indicate the homoscedastic uncertainty weighting strategy. It is immediately noticeable how it greatly improves our results. The comparison has been performed on a Titan Xp GPU, using 640\(\times\)480 images. Real performances might slightly vary as the implementation details for the top-three approaches are only partially public. The networks have been reproduced to the best of our knowledge.}
\end{table}

\subsection{Qualitative evaluation}

We then proceeded to test the system on sequences taken by the IVVI 2.0 intelligent vehicle \cite{martin2014ivvi} outside the campus of Universidad Carlos III de Madrid, located in Leganés, using our ROS implementation with post-processing enabled. The results can be visually evaluated in figure \ref{results_images}. In the provided examples, it can be viewed how the detection is well generalized even on unseen images. Also, in many images the road markings are heavily damaged, and this gives proof of the robustness of the approach.

\section{Conclusions and future work}
In conclusion, we can summarize our main contributions in the following points: we designed a drivable areas detection algorithm that takes into account different lanes, and we provided an efficient implementation of it based on the ROS framework. We used the homoscedastic uncertainty estimation to achieve better performances in the training procedure of a multi-task CNN. This training strategy made it possible to exploit image-level information, without losing accuracy on the free space detection task and facilitating a possible lane change. With our approach, we extract information from a street scene that usually requires two CNNs and a data fusion algorithm, with a single CNN, and we exploited image-level labels to define what is usable in our detection, in a novel way. Future researches will be conducted to evaluate a voting algorithm over time for a consistent prediction of the road class that will be useful to plan the vehicle path as defined in section \ref{exploiting}. In addition to this, we will investigate the possibility to modify the CNN to obtain a clustered output and reduce the impact of the post-processing algorithm on the inference times.

\bibliographystyle{IEEEtran}
\bibliography{IEEEabrv,bibliography}

\begin{thebibliography}{10}
\providecommand{\url}[1]{#1}
\csname url@rmstyle\endcsname
\providecommand{\newblock}{\relax}
\providecommand{\bibinfo}[2]{#2}
\providecommand\BIBentrySTDinterwordspacing{\spaceskip=0pt\relax}
\providecommand\BIBentryALTinterwordstretchfactor{4}
\providecommand\BIBentryALTinterwordspacing{\spaceskip=\fontdimen2\font plus
\BIBentryALTinterwordstretchfactor\fontdimen3\font minus
  \fontdimen4\font\relax}
\providecommand\BIBforeignlanguage[2]{{%
\expandafter\ifx\csname l@#1\endcsname\relax
\typeout{** WARNING: IEEEtran.bst: No hyphenation pattern has been}%
\typeout{** loaded for the language `#1'. Using the pattern for}%
\typeout{** the default language instead.}%
\else
\language=\csname l@#1\endcsname
\fi
#2}}

\bibitem{narote2018review}
S.~P. Narote, P.~N. Bhujbal, A.~S. Narote, and D.~M. Dhane, ``A review of
  recent advances in lane detection and departure warning system,''
  \emph{Pattern Recognition}, vol.~73, pp. 216--234, 2018.

\bibitem{hillel2014recent}
A.~B. Hillel, R.~Lerner, D.~Levi, and G.~Raz, ``Recent progress in road and
  lane detection: a survey,'' \emph{Machine vision and applications}, vol.~25,
  no.~3, pp. 727--745, 2014.

\bibitem{katramados2009real}
I.~Katramados, S.~Crumpler, and T.~P. Breckon, ``Real-time traversable surface
  detection by colour space fusion and temporal analysis,'' in
  \emph{International Conference on Computer Vision Systems}, 2009.

\bibitem{alvarez2007shadow}
J.~M. {\'A}lvarez, A.~M. L{\'o}pez, and R.~Baldrich, ``Shadow resistant road
  segmentation from a mobile monocular system,'' in \emph{Iberian Conference on
  Pattern Recognition and Image Analysis}.\hskip 1em plus 0.5em minus
  0.4em\relax Springer, 2007, pp. 9--16.

\bibitem{felisa2010robust}
M.~Felisa and P.~Zani, ``Robust monocular lane detection in urban
  environments,'' in \emph{Intelligent Vehicles Symposium (IV)}.\hskip 1em plus
  0.5em minus 0.4em\relax IEEE, 2010.

\bibitem{NEVEN2018T}
D.~Neven, B.~D. Brabandere, S.~Georgoulis, M.~Proesmans, and L.~V. Gool,
  ``Towards end-to-end lane detection: an instance segmentation approach,''
  \emph{2018 IEEE Intelligent Vehicles Symposium (IV)}, 2018.

\bibitem{lee2017vpgnet}
S.~Lee, J.~Kim, J.~S. Yoon, S.~Shin, O.~Bailo, N.~Kim, T.-H. Lee, H.~S. Hong,
  S.-H. Han, and I.~S. Kweon, ``Vpgnet: Vanishing point guided network for lane
  and road marking detection and recognition,'' in \emph{2017 IEEE ICCV}.\hskip
  1em plus 0.5em minus 0.4em\relax IEEE, 2017, pp. 1965--1973.

\bibitem{liu2018lane}
J.~Liu, L.~Lou, D.~Huang, Y.~Zheng, and W.~Xia, ``Lane detection based on
  straight line model and k-means clustering,'' in \emph{2018 IEEE 7th Data
  Driven Control and Learning Systems Conference (DDCLS)}.\hskip 1em plus 0.5em
  minus 0.4em\relax IEEE, 2018, pp. 527--532.

\bibitem{wang2018lane}
J.~Wang, W.~Hong, and L.~Gong, ``Lane detection algorithm based on density
  clustering and ransac,'' in \emph{2018 Chinese Control And Decision
  Conference (CCDC)}.\hskip 1em plus 0.5em minus 0.4em\relax IEEE, 2018, pp.
  919--924.

\bibitem{xu2017lane}
Y.~Xu, X.~Shan, B.~Chen, C.~Chi, Z.~Lu, and Y.~Wang, ``A lane detection method
  combined fuzzy control with ransac algorithm,'' in \emph{Power Electronics
  Systems and Applications-Smart Mobility, Power Transfer \& Security
  (PESA)}.\hskip 1em plus 0.5em minus 0.4em\relax IEEE, 2017.

\bibitem{liu2018co}
Z.~Liu, S.~Yu, and N.~Zheng, ``A co-point mapping-based approach to drivable
  area detection for self-driving cars,'' \emph{Engineering}, 2018.

\bibitem{liu2018segmentation}
X.~Liu and Z.~Deng, ``Segmentation of drivable road using deep fully
  convolutional residual network with pyramid pooling,'' \emph{Cognitive
  Computation}, vol.~10, no.~2, pp. 272--281, 2018.

\bibitem{sanberg2017free}
W.~P. Sanberg, G.~Dubbleman, \emph{et~al.}, ``Free-space detection with
  self-supervised and online trained fully convolutional networks,''
  \emph{Electronic Imaging}, vol. 2017, no.~19, pp. 54--61, 2017.

\bibitem{caltagirone2019lidar}
L.~Caltagirone, M.~Bellone, L.~Svensson, and M.~Wahde, ``Lidar--camera fusion
  for road detection using fully convolutional neural networks,''
  \emph{Robotics and Autonomous Systems}, vol. 111, 2019.

\bibitem{Teichmann2018MultiNetRJ}
M.~Teichmann, M.~Weber, J.~M. Z{\"o}llner, R.~Cipolla, and R.~Urtasun,
  ``Multinet: Real-time joint semantic reasoning for autonomous driving,''
  \emph{2018 IEEE Intelligent Vehicles Symposium (IV)}, 2018.

\bibitem{DBLP:journals/corr/abs-1806-05525}
M.~Ghafoorian, C.~Nugteren, N.~Baka, O.~Booij, and M.~Hofmann, ``El-gan:
  Embedding loss driven generative adversarial networks for lane detection,''
  \emph{CoRR}, vol. abs/1806.05525, 2018.

\bibitem{long2015fully}
J.~Long, E.~Shelhamer, and T.~Darrell, ``Fully convolutional networks for
  semantic segmentation,'' in \emph{CVPR}, 2015.

\bibitem{romera2018erfnet}
E.~Romera, J.~M. Alvarez, L.~M. Bergasa, and R.~Arroyo, ``Erfnet: Efficient
  residual factorized convnet for real-time semantic segmentation,'' \emph{IEEE
  Transactions on Intelligent Transportation Systems}, 2018.

\bibitem{cordts2016cityscapes}
M.~Cordts, M.~Omran, S.~Ramos, T.~Rehfeld, M.~Enzweiler, R.~Benenson,
  U.~Franke, S.~Roth, and B.~Schiele, ``The cityscapes dataset for semantic
  urban scene understanding,'' in \emph{CVPR}, 2016.

\bibitem{yu2018bdd100k}
F.~Yu, W.~Xian, Y.~Chen, F.~Liu, M.~Liao, V.~Madhavan, and T.~Darrell,
  ``Bdd100k: A diverse driving video database with scalable annotation
  tooling,'' \emph{CoRR}, vol. abs/1805.04687, 2018.

\bibitem{kendall2017multi}
A.~Kendall, Y.~Gal, and R.~Cipolla, ``Multi-task learning using uncertainty to
  weigh losses for scene geometry and semantics,'' in \emph{CVPR}, 2018.

\bibitem{paszke2016enet}
A.~Paszke, A.~Chaurasia, S.~Kim, and E.~Culurciello, ``Enet: A deep neural
  network architecture for real-time semantic segmentation,'' \emph{CoRR}, vol.
  abs/1606.02147, 2016.

\bibitem{pan2018two}
X.~Pan, P.~Luo, J.~Shi, and X.~Tang, ``Two at once: Enhancing learning and
  generalization capacities via ibn-net,'' in \emph{ECCV}, 2018.

\bibitem{he2016deep}
K.~He, X.~Zhang, S.~Ren, and J.~Sun, ``Deep residual learning for image
  recognition,'' in \emph{CVPR}, 2016.

\bibitem{martin2014ivvi}
D.~Mart{\'\i}n, F.~Garc{\'\i}a, B.~Musleh, D.~Olmeda, G.~Pel{\'a}ez,
  P.~Mar{\'\i}n, A.~Ponz, C.~Rodr{\'\i}guez, A.~Al-Kaff, A.~de~la Escalera,
  \emph{et~al.}, ``Ivvi 2.0: An intelligent vehicle based on computational
  perception,'' \emph{Expert Systems with Applications}, vol.~41, no.~17, pp.
  7927--7944, 2014.

\end{thebibliography}

\end{document}